%% file: acl_latex.tex
\documentclass[11pt]{article}

\usepackage[preprint]{acl}

\usepackage{latexsym}
\usepackage{microtype}
\usepackage{inconsolata}

\usepackage{graphicx}
\usepackage{tcolorbox}
\usepackage{multirow}
\usepackage{algorithm}
\usepackage{booktabs}
\usepackage{makecell}

\usepackage[english,bidi=default]{babel}

\babelfont{rm}{TeXGyreTermesX}

\babelprovide[import]{arabic}
\babelfont[arabic]{rm}[
    Path=fonts/
]{NotoSansArabic-Regular.ttf}

\title{Compression Through a Context Lens: Evaluating Adaptation, Quantization, and Distillation in Arabic Legal LLMs}

\author{
  Md. Rafiul Biswas \and George Mikros \\
  Hamad Bin Khalifa University \\
  Doha, Qatar \\
  \texttt{\{mbiswas,GMikros\}@hbku.edu.qa}
  \And
  Wajdi Zaghouani \\
  Northwestern University in Qatar \\
  Doha, Qatar \\
  \texttt{wajdi.zaghouani@northwestern.edu}
}

\begin{document}
\maketitle

\begin{abstract}
Model compression can facilitate the deployment of
large language models in resource-constrained legal
settings, but its effects on domain-specific performance
may depend on the evaluation protocol. We investigate
this issue using \textsc{ArabLegalEval}, a Saudi Arabic
legal benchmark, under both with-context and no-context
evaluation. The former provides supporting legal
passages, while the latter assesses performance without
external context. Removing the context produces
accuracy gaps of up to 27 percentage points for the base model. We examine three stages: domain adaptation,
quantization, and knowledge distillation across
FANAR-9B, JAIS-8B, and Qwen2.5-7B. Four-bit
quantization reduces model memory by 64--71\%,
with changes in no-context MCQ accuracy ranging
from $-3.2$ to $+1.6$ percentage points.
Distillation produces more variable results,
with accuracy generally declining as student
size decreases. A Qwen2.5-7B student distilled
from FANAR achieves 76.6\% no-context accuracy,
compared with 77.2\% for its teacher.
Our findings suggest that no-context
evaluation provides a useful complement to
context-supported testing when assessing the
effects of adaptation and compression on
Arabic legal language models.\footnote{https://anonymous.4open.science/r/arabic-LLM-legal-compression-6B5A}
\end{abstract}

\section{Introduction}
\label{sec:introduction}

Large language models (LLMs) have demonstrated
promising capabilities in domain-specific
applications, including legal reasoning, contract
analysis, and legal question answering
\cite{chalkidis2020legal,katz2023natural,ariai2025natural}.
However, deploying these models in legal settings
remains challenging, particularly where computational
resources and infrastructure are limited
\cite{dong2025fine,panjolia2026effective}.
Local deployment may help address confidentiality
concerns, although the use of LLMs in legal
applications still requires appropriate human
oversight, jurisdiction-specific validation, and
safeguards against inaccurate legal advice.

Recent Arabic LLMs, including FANAR-9B
\cite{fanarllm2025}, Jais-13B
\cite{anwar2026jais}, and AceGPT-13B
\cite{huang2024acegpt}, have advanced Arabic
language processing. Nevertheless, their
computational requirements can limit deployment
in resource-constrained environments.
The sensitive nature of legal information may
also motivate local inference rather than
reliance on external services
\cite{heydari2025tiny,guo2018cloud}.

Model compression offers a potential approach
to reducing these computational requirements
\citep{gou2021knowledge,xu2024survey,
alyafeai-ahmad-2021-arabic}.
However, much of the existing research has
focused on general-purpose or English-centric
benchmarks \cite{ariai2025natural}, leaving
the effects of compression on Arabic legal
tasks less explored. An additional challenge
concerns evaluation: when a supporting legal
passage is provided, models may answer questions
by using the supplied information rather than
relying on knowledge acquired during training.
Consequently, high with-context accuracy does
not necessarily indicate comparable performance
when supporting context is unavailable.

We investigate this distinction by evaluating
Arabic legal LLMs under both with-context and
no-context protocols while examining the effects
of domain adaptation, quantization, and knowledge
distillation. Our primary focus is no-context
multiple-choice question (MCQ) accuracy as an
indicator of context-independent task performance.
We address the following research questions:

\begin{itemize}
    \item \textbf{RQ1 (Context dependency).}
    How does access to supporting legal context
    affect MCQ accuracy, and how does the
    with-context/no-context gap vary across
    models and adaptation strategies?

    \item \textbf{RQ2 (Adaptation and quantization).}
    How do continued pretraining (CPT),
    supervised fine-tuning (SFT), their sequential
    combination (CPT$\to$SFT), and post-training
    quantization affect no-context MCQ accuracy
    across different model families?

    \item \textbf{RQ3 (Distillation scaling).}
    How does no-context MCQ accuracy vary with
    student size and teacher--student configuration,
    including cross-architecture distillation?
\end{itemize}

We conduct our experiments on
\textsc{ArabLegalEval}~\citep{hijazi2024arablegaleval},
an Arabic legal benchmark based on Saudi legal
sources. We evaluate three open-source models:
FANAR-9B \cite{fanarllm2025},
JAIS-8B \cite{anwar2026jais}, and
Qwen2.5-7B \cite{qwen2.5}.
First, we compare the base models with CPT,
SFT, and CPT$\to$SFT adaptation. We then
examine post-training quantization of the
merged CPT$\to$SFT checkpoints and, separately,
distillation from the SFT-only checkpoints
into smaller student models. All three stages
are evaluated using the same fixed MCQ subset
under both protocols. We additionally report
open-ended question-answering performance
on NajizQA as a secondary evaluation.

The remainder of this paper is organized as
follows. Section~2 reviews related work.
Section~3 describes the methodology, and
Section~4 presents the experimental setup.
Section~5 reports the results, followed by
the discussion in Section~6 and conclusion
in Section~7.

\section{Related Work}
\label{sec:related_work}

\subsection{Arabic LLMs and Legal-Domain Evaluation}

The Arabic LLM landscape has grown rapidly, progressing from BERT-style encoders to
instruction-tuned generative models such as Jais~\citep{anwar2026jais},
AceGPT~\citep{huang2024acegpt}, ALLaM~\citep{bari2025allam}, and
FANAR~\citep{fanarllm2025}. Nevertheless, evaluations of these models in specialized legal
domains remain limited. ArabLegalEval~\citep{hijazi2024arablegaleval} evaluates Arabic legal
understanding, while MizanQA~\citep{bahaj2026mizanqa} focuses on Moroccan law. Other benchmarks
address related areas, including Islamic inheritance reasoning in QIAS-2026
\citep{bouchekif2026qias} and MAWARITH~\citep{bouchekif2026mawarith}, broader Islamic
knowledge in \textsc{IslamicMMLU}~\citep{abdelaal2026islamicmmlu}, and financial and
Shari'ah-compliant reasoning in SAHM~\citep{sahm2026}.

To our knowledge, prior work has not systematically examined how Arabic legal knowledge changes
under model compression or how this knowledge should be evaluated when access to source context
may confound knowledge with extraction ability. We therefore use ArabLegalEval as a testbed for
studying the effects of adaptation, quantization, and distillation rather than treating benchmark
performance as an endpoint in itself.

\subsection{LLM Compression: Quantization, Distillation, and Adaptation}

Post-training quantization methods, including GPTQ~\citep{frantar2022gptq},
AWQ~\citep{lin2025awq}, and NF4~\citep{dettmers2023qlora}, reduce model size and memory
requirements by lowering weight precision. Knowledge distillation~\citep{hinton2015distilling}
transfers capabilities from a larger teacher to a smaller student, while parameter-efficient
methods such as LoRA~\citep{hu2022lora} and QLoRA~\citep{dettmers2023qlora} support domain
adaptation with limited trainable parameters. Recent studies have examined these approaches
within broader efficiency frameworks~\citep{von2026unicomp,zhu2024survey}. 

Much of the existing evaluation literature focuses on English or multilingual benchmarks, where the accuracy--size trade-off may differ from morphologically rich, lower-resource languages such as Arabic. Closest to our work,
\citet{alshehhi2025inclusive} compress multilingual transformers on Arabic, English, and
Kannada, finding that $4$- and $8$-bit quantization reduce memory with minimal accuracy loss
on ArabicMMLU~\citep{koto2024arabicmmlu}. We differ in three ways: our domain is Arabic
\emph{legal} understanding; we evaluate adaptation, quantization, and distillation as
three-stage experimental framework on shared models and benchmark; and, most importantly, we
evaluate every stage under both with-context and no-context protocols, which reframes what
``preserved performance'' under compression actually means.

\section{Method}
\label{sec:method}

\begin{figure*}[t]
    \centering
    \includegraphics[width=\textwidth]{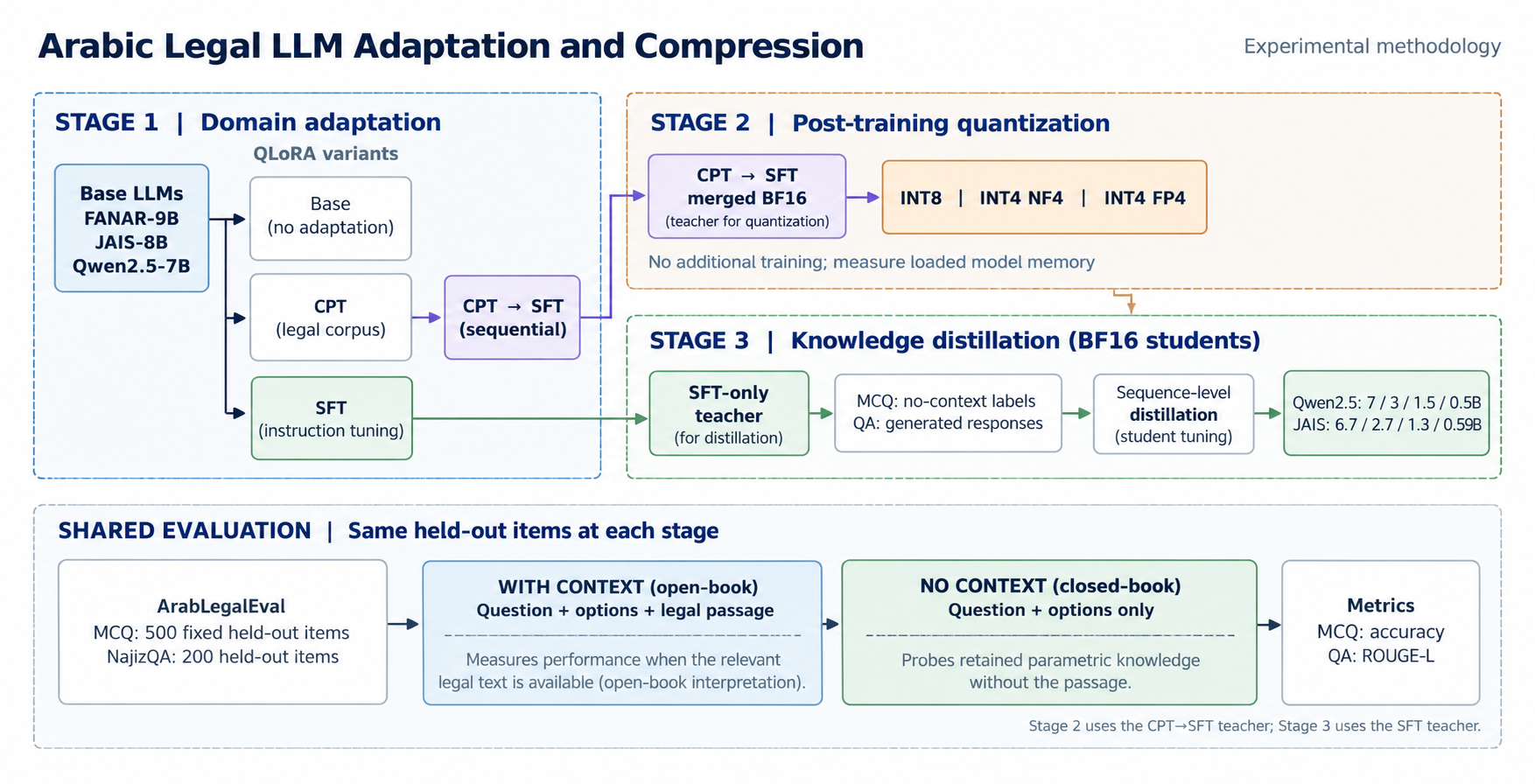}
    \caption{Overview of the proposed methodology.}
    \label{fig:methodology}
\end{figure*}

We examine how adaptation and compression affect
Arabic legal question-answering performance through
a three-stage experimental framework: adaptation,
quantization, and distillation. The framework is
applied to three base LLMs, with each stage
evaluated under two protocols
(Figure~\ref{fig:methodology}).
This section describes the benchmark and data
splits (Section~\ref{sec:method_setup}),
the dual-protocol evaluation
(Section~\ref{sec:method_eval}),
and the three experimental stages
(Sections~\ref{sec:method_adapt}--\ref{sec:method_distill}).

\subsection{Benchmark, Data, and Base Models}
\label{sec:method_setup}

We use \textsc{ArabLegalEval}~\citep{hijazi2024arablegaleval}, an Arabic legal
benchmark constructed from Saudi Ministry of Justice and Board of Experts documents. For MCQ, we use a fixed 80/20 train--evaluation split with seed $42$, giving $2{,}113$ held-out examples. For NajizQA, we sample up to $2{,}000$ examples and apply a 90/10 split, yielding $200$ evaluation items. All adaptation experiments in Section~\ref{sec:method_adapt} use only the corresponding training splits.

To keep evaluation consistent across the full compression pipeline, we report MCQ performance on a fixed subset of $500$ held-out examples and NajizQA performance on all $200$ held-out examples. The same evaluation sets are used for adaptation, quantization, and distillation, so results remain directly comparable across stages.

We study three open-source instruction-tuned LLMs: the Arabic-centric \textbf{FANAR-9B}~\citep{fanarllm2025} and \textbf{JAIS-8B}~\citep{anwar2026jais}, and the multilingual \textbf{Qwen2.5-7B-Instruct}~\citep{qwen2.5}. This mix allows us to examine whether the effects of compression and context dependence differ between Arabic-specialized and general multilingual models. Each model is processed independently through the full three-stage pipeline.

\paragraph{Data separation.}
Training, checkpoint selection, and hyperparameter
tuning were conducted without using the held-out
evaluation examples. We additionally checked
for overlap between the training and evaluation
sets and found no overlap under our verification
procedure. The fixed 500-item MCQ subset was
reserved for final evaluation.

\subsection{Dual-Protocol Evaluation}
\label{sec:method_eval}

A single MCQ score on \textsc{ArabLegalEval}
may reflect both the use of supplied legal
context and knowledge acquired during training.
To examine answer availability in the supplied
passages, we checked whether the correct option
text appeared as an exact substring of the legal
context. Among the 500 evaluation examples,
270 (54.0\%) contained the correct answer text.
This measure captures direct textual overlap
but does not establish whether a model relies
on that overlap when generating its prediction (see in appendix Section Additional Results).

\begin{itemize}

\item \textbf{With-context (open-book).}
The prompt includes the question, four options,
and the associated legal context. Performance
in this setting reflects the model's ability
to answer questions with access to supporting
information, including its use of the supplied
passage.

\item \textbf{No-context (closed-book).}
The legal context is removed, and the model
receives only the question and answer options.
We use this setting as the primary measure
of context-independent legal MCQ performance,
which may partly reflect knowledge acquired
during pretraining and adaptation.

\end{itemize}

For MCQ, models are prompted in Arabic to return a single option label (A--D). We extract the first valid Latin or Arabic option label from the generated response and compute exact-match accuracy. For NajizQA, we report ROUGE-L against the reference answer. Decoding settings are fixed across models and stages, with temperature $0.1$ and top-$p$ $0.9$.

\subsection{Stage 1: Adaptation}
\label{sec:method_adapt}

The first stage compares adaptation strategies
for each base model. We evaluate the unadapted
model against three strategies, implemented
using parameter-efficient QLoRA fine-tuning
\citep{hu2022lora,dettmers2023qlora}
(a LoRA adapter over a 4-bit NF4-quantized
backbone):

\begin{itemize}
    \item \textbf{Base.} The instruction-tuned model evaluated zero-shot, with no legal
    adaptation.
    \item \textbf{Continual pre-training (CPT).} Causal language-model training on a
    legal corpus assembled from the training-split MCQ and QA items, each linearized
    into an Arabic template of context, question, correct answer, and explanation.
    \item \textbf{Supervised fine-tuning (SFT).} Instruction-style fine-tuning on the
    same training items formatted as (instruction, response) pairs, where the
    instruction presents the legal question (and options for MCQ) and the response is the
    correct answer and explanation.
    \item \textbf{Sequential combination (CPT$\to$SFT).} SFT applied on top of the CPT
    checkpoint, combining domain-knowledge injection with task-format adaptation.
\end{itemize}

Each of the four variants is evaluated under both protocols
(Section~\ref{sec:method_eval}). 
The CPT$\to$SFT checkpoint of each model
is used for Stage~2 quantization, while
the corresponding SFT-only checkpoint
serves as the teacher for Stage~3
distillation. These choices allow the two
compression methods to be examined under
different adaptation settings, although
their absolute results are not directly
comparable as effects of compression alone. 
The adaptation recipe and training
hyperparameters are held constant across
all three base models
(Section~\ref{sec:setup}), allowing us
to compare their responses to the same
adaptation strategies. Differences in
performance may nevertheless reflect
model-specific characteristics and
interactions with the shared training
configuration.

\subsection{Stage 2: Post-Training Quantization}
\label{sec:method_quant}

In the second stage, we compress the selected Stage~1 teacher using post-training quantization,
without any additional training. For each base model, the selected CPT$\to$SFT adapter is merged
into the base weights to produce a single BF16 checkpoint, which serves as the uncompressed
reference for quantization. (Distillation in Stage~3 instead starts from the SFT checkpoint; see
Section~\ref{sec:method_distill}.)

We evaluate three \texttt{bitsandbytes} quantization settings: INT8 using the outlier-aware
\texttt{LLM.int8()} method, INT4~NF4 using NormalFloat quantization with double quantization and
BF16 computation~\citep{dettmers2023qlora}, and INT4~FP4 with the same 4-bit configuration but a
floating-point representation. The two 4-bit variants let us compare whether the choice of
low-bit number format affects legal-task performance.

For each quantized model, we record its memory footprint, compression relative to the BF16
teacher, and MCQ accuracy under both with-context and no-context evaluation. This comparison
shows whether quantization mainly affects context-based answer extraction or also degrades the
legal knowledge retained in the model parameters.

\subsection{Stage 3: Knowledge Distillation}
\label{sec:method_distill}

In the third stage, we transfer the teacher's behavior to smaller student models. Unlike Stage~2, which uses the teacher selected for quantization, distillation uses the Stage~1 \textbf{SFT} checkpoint. We first generate pseudo-labels over the training split. For MCQ items, the teacher is queried in the no-context setting so that the supervision reflects retained legal knowledge rather than direct extraction from the context. 

Each student is then fine-tuned on these prompt--response pairs with a standard causal language-modeling objective. This corresponds to sequence-level, hard-label distillation~\citep{kim-rush-2016-sequence}, where the student learns from the teacher's generated outputs rather than its full probability distribution. We use this formulation because it is simple and does not require the teacher and student to share the same tokenizer or vocabulary, which also makes cross-family distillation possible. Logit-level distillation with temperature scaling is left for future work.

\section{Experimental Setup}
\label{sec:setup}

We conduct experiments on \textsc{ArabLegalEval}
\citep{hijazi2024arablegaleval}, focusing on its
multiple-choice questions (MCQ) and open-ended
question answering (NajizQA) tasks. Using a fixed
random seed, we split the MCQ data into 8,451
training and 2,113 evaluation examples, and
NajizQA into 1,800 training and 200 evaluation
examples. Adaptation and distillation use only
the training data. For evaluation, we use a
fixed subset of 500 held-out MCQ examples and
all 200 NajizQA examples across the relevant
experiments. Further details on MCQ subsampling
are provided in Appendix A.

We evaluate three instruction-tuned models:
\textbf{FANAR-9B}~\citep{fanarllm2025},
\textbf{JAIS-8B}~\citep{anwar2026jais},
and \textbf{Qwen2.5-7B}~\citep{qwen2.5}.
Stage~1 performs domain adaptation using
QLoRA~\citep{hu2022lora,dettmers2023qlora}.
Stage~2 applies INT8, INT4~NF4, and INT4~FP4
quantization to the merged CPT$\to$SFT
checkpoints through \texttt{bitsandbytes}.
Stage~3 separately distills the SFT-only
teachers into Qwen2.5 or JAIS student models,
depending on the teacher family.

All models are evaluated under with-context
and no-context protocols. We report exact-match
MCQ accuracy based on the predicted option
label and ROUGE-L for NajizQA. The primary
evaluation focuses on no-context MCQ accuracy,
while NajizQA provides a complementary measure
of open-ended answer generation. We also examine
answer-in-context overlap by checking whether
the correct option text appears verbatim in
the supplied legal passage.

Experiments are conducted on an HPC system
equipped with NVIDIA A100 GPUs and 48\,GB
of system RAM. Individual Stage~2 and Stage~3
experiments take approximately 16-18h hours.
Additional training hyperparameters, prompts,
decoding settings, and implementation details
are provided in Appendix A.

\section{Results}
\label{sec:results}

We present the results across three pipeline stages:
domain adaptation (Section~\ref{sec:results_stage1}),
quantization (Section~\ref{sec:results_stage2}), and
knowledge distillation (Section~\ref{sec:results_stage3}).
Each stage is evaluated under two protocols: with
supporting context and without context. Unless otherwise
specified, MCQ accuracy is reported on a fixed
500-item evaluation subset, while open-ended QA is
evaluated on 200 items (Section~\ref{sec:setup}).
Additional results are provided in
Appendix A.

\subsection{Stage 1: Adaptation and the Context Gap}
\label{sec:results_stage1}

Stage~1 investigates RQ1 and the adaptation component
of RQ2. Table~\ref{tab:s1} compares the base models
with three adaptation strategies: continued
pretraining (CPT), supervised fine-tuning (SFT),
and their sequential combination (CPT$\to$SFT).

\paragraph{Context gap.}
All three base models perform better when supporting
context is available. FANAR exhibits the largest
difference, dropping from 97.8\% with context to
70.8\% without context (27.0 percentage points),
followed by Qwen2.5-7B (21.4 points) and JAIS-8B
(7.6 points). These gaps indicate that high
context-supported accuracy does not necessarily
translate into comparable performance when models
must rely on parametric knowledge. Notably,
CPT produces contrasting effects for FANAR:
no-context accuracy improves by 4.0 points,
while with-context accuracy declines by 55.2 points.

\paragraph{Effect of adaptation.}
Adaptation yields model-dependent improvements in
no-context MCQ accuracy. FANAR benefits most from
the combined CPT$\to$SFT strategy, improving from
70.8\% to 79.2\% (+8.4 points). Qwen2.5-7B
achieves a smaller gain, increasing from 76.0\%
to 79.8\% (+3.8 points). JAIS-8B responds
differently: CPT reduces no-context accuracy
from 71.6\% to 61.2\%, while SFT alone achieves
the highest result (76.4\%), slightly exceeding
CPT$\to$SFT (75.8\%). Thus, sequential adaptation
does not consistently outperform SFT, and the
benefits of domain adaptation vary across
model families and training objectives.

\paragraph{Open-ended QA.}
Performance on NajizQA reveals a different pattern
from MCQ evaluation. Despite improved MCQ accuracy,
JAIS obtains near-zero ROUGE-L scores after SFT
and CPT$\to$SFT. FANAR achieves approximately
0.17 ROUGE-L with context after adaptation, but
its SFT-based variants perform considerably worse
without context (0.0490--0.0637). Qwen2.5-7B
achieves the highest ROUGE-L scores following
SFT (0.3170 with context and 0.2950 without
context), exceeding its CPT$\to$SFT variant.
These differences indicate that improvements
in structured answer selection do not necessarily
transfer to open-ended generation.

\begin{table*}[t]
\centering
\small
\setlength{\tabcolsep}{6pt}
\begin{tabular}{llccccc}
\toprule
& & \multicolumn{3}{c}{\textbf{MCQ Accuracy (\%)}} & \multicolumn{2}{c}{\textbf{QA ROUGE-L}} \\
\cmidrule(lr){3-5}\cmidrule(lr){6-7}
\textbf{Model} & \textbf{Variant} & \textbf{With-Ctx} & \textbf{No-Ctx} & \textbf{Gap ($\Delta$)} & \textbf{With-Ctx} & \textbf{No-Ctx} \\
\midrule
\multirow{4}{*}{FANAR-9B}
 & Base        & 97.8 & 70.8 & $+27.0$ & 0.0083 & 0.0058 \\
 & CPT         & 42.6 & 74.8 & $-32.2$ & 0.1700 & 0.1450 \\
 & SFT         & \textbf{98.8} & 77.2 & $+21.6$ & 0.1700 & 0.0490 \\
 & CPT$\to$SFT & \textbf{98.8} & \textbf{79.2} & $+19.6$ & 0.1683 & 0.0637 \\
\midrule
\multirow{4}{*}{JAIS-8B}
 & Base        & 79.2 & 71.6 & $+7.6$  & 0.0194 & 0.0179 \\
 & CPT         & 93.6 & 61.2 & $+32.4$ & 0.0249 & 0.0008 \\
 & SFT         & 95.8 & \textbf{76.4} & $+19.4$ & 0.0001 & 0.0009 \\
 & CPT$\to$SFT & \textbf{98.6} & 75.8 & $+22.8$ & 0.0012 & 0.0012 \\
\midrule
\multirow{4}{*}{Qwen2.5-7B}
 & Base        & 97.4 & 76.0 & $+21.4$ & 0.0095 & 0.0058 \\
 & CPT         & 97.6 & 78.4 & $+19.2$ & 0.1532 & 0.2152 \\
 & SFT         & 98.6 & 78.2 & $+20.4$ & 0.3170 & 0.2950 \\
 & CPT$\to$SFT & \textbf{98.8} & \textbf{79.8} & $+19.0$ & 0.2750 & 0.2637 \\
\bottomrule
\end{tabular}
\caption{Stage~1 results for three base models and four adaptation variants. MCQ accuracy is reported on the fixed 500-item set and QA ROUGE-L on 200 items; $\Delta$ denotes With-Ctx minus No-Ctx. Bold marks the best variant per model and protocol (ties both bolded). Raw counts and ROUGE-1 are given in Appendix Table 4}
\label{tab:s1}
\end{table*}

\begin{table*}[t]
\centering
\small
\setlength{\tabcolsep}{6pt}
\begin{tabular}{llccccc}
\toprule
& & & & \multicolumn{2}{c}{\textbf{MCQ Accuracy (\%)}} & \\
\cmidrule(lr){5-6}
\textbf{Model} & \textbf{Method} & \textbf{Mem.\ (GB)} & \textbf{Compr.\ (\%)} & \textbf{With-Ctx} & \textbf{No-Ctx} & \textbf{$\Delta$No-Ctx} \\
\midrule
\multirow{4}{*}{FANAR-9B}
 & BF16 (Baseline) & 16.36 & --- & 98.8 & 79.2 & --- \\
 & INT8            & 8.61  & 47.4 & 99.2 & 79.4 & $+0.2$ \\
 & INT4 NF4        & 4.73  & 71.1 & 99.0 & 79.4 & $+0.2$ \\
 & INT4 FP4        & 4.73  & 71.1 & 98.2 & 79.2 & $0.0$ \\
\midrule
\multirow{4}{*}{JAIS-8B}
 & BF16 (Baseline) & 15.07 & --- & 98.6 & 75.8 & --- \\
 & INT8            & 8.47  & 43.8 & 98.2 & 75.8 & $0.0$ \\
 & INT4 NF4        & 5.17  & 65.7 & 97.4 & 77.4 & $+1.6$ \\
 & INT4 FP4        & 5.17  & 65.7 & 98.0 & 74.2 & $-1.6$ \\
\midrule
\multirow{4}{*}{Qwen2.5-7B}
 & BF16 (Baseline) & 14.19 & --- & 98.8 & 79.8 & --- \\
 & INT8            & 8.11  & 42.8 & 98.8 & 77.8 & $-2.0$ \\
 & INT4 NF4        & 5.07  & 64.3 & 98.6 & 77.8 & $-2.0$ \\
 & INT4 FP4        & 5.07  & 64.3 & 99.0 & 76.6 & $-3.2$ \\
\bottomrule
\end{tabular}

\caption{Stage~2 quantization results on the fixed
500-item MCQ evaluation set. BF16 baselines are
the CPT$\to$SFT models from Table~\ref{tab:s1}.
\textbf{Compr.} denotes the percentage reduction
in model memory relative to BF16, and
\textbf{$\Delta$No-Ctx} indicates the change
in no-context accuracy in percentage points.}

\label{tab:s2}
\end{table*}




\subsection{Stage 2: Quantization}
\label{sec:results_stage2}

\paragraph{Impact on no-context accuracy.}
Table~\ref{tab:s2} reports quantization results
relative to the BF16 CPT$\to$SFT models from Stage~1.
INT4 quantization reduces model memory by
64.3--71.1\%, with relatively small changes in
no-context MCQ accuracy. Under NF4, FANAR improves
by 0.2 percentage points, JAIS by 1.6 points,
while Qwen2.5-7B declines by 2.0 points.
The largest absolute change across all quantization
configurations is a 3.2-point reduction for Qwen
under INT4 FP4. These results suggest that
substantial memory savings can be achieved while
largely maintaining no-context MCQ performance
under the evaluated settings.

\paragraph{Comparison of 4-bit methods.}
NF4 achieves higher no-context accuracy than FP4
across all three model families. The largest
difference occurs for JAIS (77.4\% versus 74.2\%),
followed by Qwen (77.8\% versus 76.6\%) and FANAR
(79.4\% versus 79.2\%). Although these differences
are modest, NF4 consistently yields the better
observed results without additional memory cost.

\paragraph{With-context performance.}
With-context accuracy remains high across all
quantized configurations, ranging from 97.4\%
to 99.2\%. This near-ceiling performance limits
the ability of the with-context protocol to
distinguish between quantization methods.
In contrast, the no-context evaluation reveals
larger differences and provides a more informative
measure of performance when external context is
unavailable.

\subsection{Stage 3: Distillation}
\label{sec:results_stage3}

Stage~3 examines how no-context MCQ accuracy varies
with student size and teacher architecture
(Table~\ref{tab:s3}). Each Stage~1 SFT teacher is
distilled into four student configurations. Qwen and
JAIS use same-family students, whereas FANAR is
distilled into Qwen2.5 models, enabling a
cross-architecture comparison. Unlike Stage~2,
the teachers are SFT-only rather than CPT$\to$SFT
checkpoints.

\paragraph{Effect of student size.}
No-context accuracy generally declines with
student size, although the magnitude varies
across teacher--student configurations.
FANAR-distilled Qwen students decrease from
76.6\% at 7B to 52.8\% at 0.5B. JAIS students
decline from 74.4\% at 6.7B to 47.6\% at 590M,
while Qwen-distilled students decrease from
76.4\% at 7B to 36.4\% at 0.5B. These results
suggest that performance retention depends
on student capacity and the distillation setup.

\paragraph{Cross-architecture distillation.}
The FANAR-distilled Qwen2.5-7B student achieves
76.6\% no-context accuracy, compared with
77.2\% for its teacher, a difference of
0.6 percentage points. The same student
architecture distilled from Qwen2.5-7B
achieves 76.4\%. These results suggest that
cross-architecture distillation can achieve
performance comparable to the teacher in
this configuration, although broader conclusions
require further evaluation.

\paragraph{Accuracy--memory trade-off.}

The smaller student has a substantially lower
estimated BF16 weight-storage requirement,
although the memory figures are not directly
equivalent because the teacher values were
measured and student values estimated. For example, the FANAR-distilled
Qwen2.5-3B student achieves 73.6\% accuracy
with an estimated BF16 weight memory of
5.76\,GB, compared with 77.2\% and 16.36\,GB
for its teacher. This represents a 3.6-point
accuracy reduction alongside substantially
lower estimated weight storage, illustrating
the trade-off between model size and
context-independent performance.

\begin{table*}[t]
\centering
\small
\setlength{\tabcolsep}{6pt}
\renewcommand{\arraystretch}{1.08}
\begin{tabular}{@{}llcccc@{}}
\toprule
& & & &
\multicolumn{2}{c}{\textbf{MCQ Accuracy (\%)}} \\
\cmidrule(lr){5-6}
\textbf{Teacher} & \textbf{Model} & \textbf{Params}
& \textbf{BF16 Mem.\ (GB)}
& \textbf{With-Ctx} & \textbf{No-Ctx} \\
\midrule
\multirow{5}{*}{FANAR-9B}
 & Teacher (SFT) & 8.7B  & 16.36 & 98.8 & \textbf{77.2} \\
 & Qwen2.5-7B    & 7.62B & $\approx$14.19 & 97.8 & 76.6 \\
 & Qwen2.5-3B    & 3.09B & $\approx$5.76  & 87.4 & 73.6 \\
 & Qwen2.5-1.5B  & 1.54B & $\approx$2.87  & 81.4 & 59.4 \\
 & Qwen2.5-0.5B  & 0.49B & $\approx$0.92  & 53.2 & 52.8 \\
\midrule
\multirow{5}{*}{JAIS-8B}
 & Teacher (SFT) & 8.0B  & 15.07 & 95.8 & \textbf{76.4} \\
 & JAIS-6.7B     & 6.7B  & $\approx$12.48 & 96.8 & 74.4 \\
 & JAIS-2.7B     & 2.7B  & $\approx$5.03  & 94.4 & 70.2 \\
 & JAIS-1.3B     & 1.3B  & $\approx$2.42  & 90.8 & 66.4 \\
 & JAIS-590M     & 0.59B & $\approx$1.10  & 72.4 & 47.6 \\
\midrule
\multirow{5}{*}{Qwen2.5-7B}
 & Teacher (SFT) & 7.62B & 14.19 & 98.6 & \textbf{78.2} \\
 & Qwen2.5-7B    & 7.62B & $\approx$14.19 & 96.2 & 76.4 \\
 & Qwen2.5-3B    & 3.09B & $\approx$5.76  & 73.6 & 66.6 \\
 & Qwen2.5-1.5B  & 1.54B & $\approx$2.87  & 66.4 & 63.2 \\
 & Qwen2.5-0.5B  & 0.49B & $\approx$0.92  & 38.8 & 36.4 \\
\bottomrule
\end{tabular}
\caption{Stage~3 distillation results on the fixed
500-item MCQ evaluation set. Teachers are the Stage~1
SFT checkpoints (Table~\ref{tab:s1}). Qwen and JAIS
use same-family students, whereas FANAR is distilled
into Qwen2.5 students. Student BF16 memory values
marked $\approx$ are weights-only estimates;
teacher memory values are measured.}
\label{tab:s3}
\end{table*}

\section{Discussion}
\label{sec:discussion}

Our findings offer several observations on evaluating
and compressing knowledge-intensive language models.

\paragraph{The importance of no-context evaluation.}
Across the three stages, with-context MCQ accuracy
was generally high, while no-context performance
showed greater variation across models and
configurations. This suggests that access to
supporting passages may reduce the visibility of
differences in context-independent performance,
as models can rely on information provided in the
input~\citep{hagstrom-etal-2026-cub}.
No-context evaluation therefore provides a useful
complement to with-context testing when examining
the effects of adaptation and compression on
retained task performance.

\paragraph{Quantization and distillation involve different trade-offs.}
INT4 quantization reduced model memory by
64--71\%, with no-context accuracy changes ranging
from $-3.2$ to $+1.6$ percentage points.
These results are consistent with previous findings
on the feasibility of low-bit model representations
\citep{dettmers2023qlora}. Distillation showed
greater variation across student sizes and model
families, with smaller students generally achieving
lower no-context accuracy. In the evaluated setting,
quantization provided substantial memory reductions
with relatively limited changes in MCQ performance,
whereas distillation offered smaller model sizes
with more variable accuracy.

\paragraph{Teacher--student configuration.}
Distillation outcomes varied across
teacher--student configurations.
The FANAR-distilled Qwen2.5-7B student
achieved 76.6\% no-context accuracy,
compared with 77.2\% for its teacher.
The same student architecture distilled
from Qwen2.5-7B achieved 76.4\%.
These observations suggest that
cross-architecture distillation is
feasible in the evaluated setting,
although the small differences do not
establish an advantage for either
teacher. Further experiments would be
needed to isolate the effects of
teacher choice and model architecture.

\section{Conclusion}
\label{sec:conclusion}

We investigated how domain adaptation, quantization,
and knowledge distillation affect Arabic legal MCQ
performance on \textsc{ArabLegalEval} under with-context
and no-context evaluation. The results show that
with-context accuracy is generally high, whereas
no-context performance varies considerably across
models and configurations. This highlights the value
of evaluating models without supporting context when
examining context-independent task performance.

The three stages exhibited different patterns.
Domain adaptation provided model-dependent improvements
in no-context accuracy, while INT4 quantization reduced
model memory by 64--71\% with relatively small observed
accuracy changes. Distillation produced more variable
results, with performance generally declining for
smaller students. Cross-architecture distillation was
feasible but did not consistently preserve teacher
performance, suggesting that teacher--student
configuration warrants further investigation.

These findings support evaluating both
context-supported and context-independent performance
when assessing compressed language models. The study
is limited to one Arabic legal benchmark, a fixed
500-item MCQ evaluation subset, and hard-label
distillation. Future work could extend the evaluation
to additional datasets, larger test sets, and
alternative distillation strategies.

\section*{Limitations}
Our study is limited to \textsc{ArabLegalEval}, Saudi law, and Arabic, so the observed context gap and compression behavior may not generalize to other jurisdictions or domains. 
MCQ results are evaluated on a fixed
500-item subset of 2,113 held-out items.
For a single accuracy estimate, the
worst-case approximate 95\% binomial
confidence-interval half-width is
4.4 percentage points. This interval
does not characterize paired differences
between models, for which item-level
paired analysis would be required. NajizQA evaluation is also limited to ROUGE-L on 200 examples, which does not directly measure legal correctness or faithfulness.

Methodologically, we use sequence-level hard-label distillation and do not evaluate soft-label distillation, pruning, low-rank factorization, or quantization-aware training. Our analysis concerns whether the evaluation prompt exposes answer-bearing context. We verified that our fine-tuning splits do not overlap with the evaluation set, but we do not test whether evaluation items appeared in the base models' pretraining data.

The distillation setup is not fully uniform across model families. JAIS uses same-family students, while FANAR is distilled into Qwen2.5 students, making teacher identity and architectural mismatch partially confounded. Student sizes are also only approximately matched across families, and JAIS memory is estimated from parameter counts rather than measured directly. In addition, Stage~3 uses the SFT teacher rather than the CPT$\to$SFT teacher used in Stage~2, so absolute distillation results may differ under an alternative teacher choice.

Future work should extend the closed-book protocol to additional benchmarks and jurisdictions, use more uniform distillation designs, and study retrieval-augmented settings where parametric and retrieved knowledge interact.

\section*{Ethical Consideration}
This study uses an existing Arabic legal benchmark and publicly available language models, and does not involve human participants or personal data. The compressed models are evaluated for research purposes and are not intended to provide legal advice. As model outputs may contain errors or reflect limitations in the underlying models and benchmark, deployment in real-world legal settings should include appropriate expert oversight and validation.

\section*{AI Assistance Declaration}
The authors used Claude Opus 5.5 and GPT-5.5 Sol to polish and improve the language of the manuscript draft, including grammar, clarity, and phrasing. The authors reviewed and approved all AI-assisted revisions and take full responsibility for the final content of the manuscript.

\bibliography{custom}
\appendix
\input{appendix}

\end{document}

%% file: appendix.tex
\appendix
\label{appendix}
\section{Implementation Details}
\label{app:implementation}

This appendix provides the implementation details summarized in Section~\ref{sec:setup}, including the data splits, evaluation subset, prompts, training configuration, decoding, and worked examples.

\subsection{Data}
\label{app:data}

We split each \textsc{ArabLegalEval} task using seed $42$. MCQ uses an $80/20$ train--evaluation split, giving $8{,}451$ training and $2{,}113$ evaluation items. For QA, we first sample at most $2{,}000$ examples and then apply a $90/10$ split, yielding $1{,}800$ training and $200$ evaluation items. Adaptation and distillation use only the training splits.

All MCQ results are reported on the same fixed $500$-item subset of the evaluation split, while QA uses all $200$ evaluation examples. The base models are \texttt{QCRI/Fanar-1-9B-Instruct}~\citep{fanarllm2025}, \texttt{inceptionai/Jais-2-8B-Chat}~\citep{anwar2026jais}, and \texttt{Qwen/Qwen2.5-7B-Instruct}~\citep{qwen2.5}.

\subsection{Evaluation Subsample}
\label{app:subsample}

The $500$ MCQ examples are sampled once from the $2{,}113$-item evaluation split using seed $42$ and kept fixed across all models, stages, and evaluation settings. This reduces evaluation cost while keeping comparisons consistent. At $n=500$, the worst-case $95\%$ Wald confidence half-width is approximately $4.4$ points, and is smaller at the accuracy levels observed in most experiments. 


\subsection{Prompts and Corpus Templates}
\label{app:prompts}

All prompts are written in Modern Standard Arabic. The two evaluation protocols use the same question, answer options, and answer cue. They differ only in the instruction and whether the legal passage is provided. In the with-context setting, the model is asked to answer using the reference legal text; in the no-context setting, the passage is removed and the model is asked to answer from its knowledge of Saudi regulations.

\begin{figure}[t]
\centering
\small

\begin{tcolorbox}[
    title={With-context MCQ prompt (gloss below)}
]
\foreignlanguage{arabic}{
أنت خبير في القانون السعودي. أجب عن السؤال التالي بناءً على النص القانوني المرجعي.
}

\medskip

\noindent
\foreignlanguage{arabic}{النص القانوني المرجعي}: \texttt{\{context\}}

\medskip

\noindent
\foreignlanguage{arabic}{السؤال}: \texttt{\{question\}}

\medskip

\noindent
\foreignlanguage{arabic}{الخيارات}:

\noindent
A) \texttt{\{A\}} \quad
B) \texttt{\{B\}} \quad
C) \texttt{\{C\}} \quad
D) \texttt{\{D\}}

\medskip

\noindent
\foreignlanguage{arabic}{
اختر الإجابة الصحيحة} (A OR B OR C OR D):

\noindent
\foreignlanguage{arabic}{الإجابة}:

\medskip

\noindent
\emph{Gloss:} ``You are an expert in Saudi law. Answer the following
question based on the reference legal text. Reference legal text:
\texttt{\{context\}}. Question: \texttt{\{question\}}. Options:
\ldots{} Choose the correct answer (A, B, C, or D only). Answer:''
\end{tcolorbox}

\medskip

\begin{tcolorbox}[
    title={No-context MCQ prompt (gloss below)}
]
\foreignlanguage{arabic}{
أنت خبير في القانون السعودي. أجب عن السؤال التالي بناءً على معرفتك بالأنظمة السعودية.
}

\medskip

\noindent
\foreignlanguage{arabic}{السؤال}: \texttt{\{question\}}

\medskip

\noindent
\foreignlanguage{arabic}{الخيارات}:

\noindent
A) \texttt{\{A\}} \quad
B) \texttt{\{B\}} \quad
C) \texttt{\{C\}} \quad
D) \texttt{\{D\}}

\medskip

\noindent
\foreignlanguage{arabic}{
اختر الإجابة الصحيحة} (A OR B OR C OR D):

\noindent
\foreignlanguage{arabic}{الإجابة}:

\medskip

\noindent
\emph{Gloss:} ``You are an expert in Saudi law. Answer the following
question based on your knowledge of Saudi regulations. Question:
\texttt{\{question\}}. Options: \ldots{} Choose the correct answer
(A, B, C, or D only). Answer:''
\end{tcolorbox}

\caption{With-context and no-context MCQ prompts. The prompts are
identical except for the instruction and the inclusion of the
\texttt{\{context\}} context in the with-context setting.}
\label{fig:prompts}
\end{figure}

\paragraph{Adaptation corpora.}
The CPT corpus linearizes each training item into an Arabic template of context (truncated
to the leading $500$ characters), question, correct answer, and explanation. The SFT set
formats the same items as (instruction, response) pairs, where the instruction presents the
question (and options, for MCQ) and the response is the correct answer plus explanation.



\subsection{Training}
\label{app:training}

\paragraph{Adaptation (Stage 1).}
All adaptation uses QLoRA~\citep{hu2022lora,dettmers2023qlora}, with LoRA adapters
($r{=}16$, $\alpha{=}32$, dropout $0.05$) applied to the attention and MLP projection layers
over a $4$-bit NF4 backbone. We train for $3$ epochs with per-device batch size $4$ and gradient
accumulation $4$ (effective batch size $16$), learning rate $2{\times}10^{-4}$, cosine decay,
$10\%$ warmup, maximum sequence length $1024$, bf16 computation, and paged 8-bit AdamW.
The training recipe is kept fixed across the three base models, apart from model-specific
checkpoints and tokenizers.

\paragraph{Quantization (Stage 2).}
For each model, the selected CPT$\to$SFT adapter is merged into the base weights to obtain a
BF16 teacher. We evaluate three \texttt{bitsandbytes} configurations: INT8 using
\texttt{LLM.int8()} with outlier threshold $6.0$, INT4~NF4 with double quantization and BF16
computation, and INT4~FP4 with the same configuration but FP4 weights.

\paragraph{Distillation (Stage 3).}
Stage~3 uses the Stage~1 SFT checkpoint as the teacher for pseudo-label generation. For MCQ,
the teacher is queried in the no-context setting and its predicted option letter is used as the
training target; predictions that cannot be parsed fall back to the gold label. For QA, the
teacher-generated response is retained. Students are trained on these prompt--response pairs
with a causal language-modeling objective, corresponding to sequence-level hard-label
distillation~\citep{kim-rush-2016-sequence}.

Qwen and FANAR teachers are distilled into Qwen2.5 students at $7$B, $3$B, $1.5$B, and
$0.5$B, while the JAIS teacher is distilled into JAIS students at $6.7$B, $2.7$B, $1.3$B,
and $590$M. Training uses the same LoRA configuration as Stage~1, with per-device batch size
$2$ and gradient accumulation $8$ (effective batch size $16$) for $3$ epochs.

\subsection{Evaluation and Decoding}
\label{app:eval}

For MCQ, we score exact match over a single predicted option label. The parser first searches
the first $50$ generated characters for one of \{A,B,C,D\}; if none is found, it maps Arabic
option labels (أ$\to$A, ب$\to$B, ج$\to$C, د$\to$D) within the first $20$ characters.
Responses with no valid option are marked \textsc{invalid} and counted as incorrect. Arabic
text is normalized by unifying common alif and ya variants, removing diacritics, and collapsing
whitespace.

Generation uses at most $10$ new tokens, temperature $0.1$, top-$p$ $0.9$, and
\texttt{do\_sample=False}, with inputs truncated to $1{,}500$ tokens. For NajizQA, we report
ROUGE-L against the reference answer. Decoding settings are fixed across models and stages.

\paragraph{Compute.}
Experiments use PyTorch~$2.12.1$ (CUDA~$13.0$), Hugging Face Transformers~$5.12.1$, PEFT,
and \texttt{bitsandbytes} on NVIDIA A100 80\,GB GPUs. We use seed $42$ throughout.
Approximate wall-clock time per model is $8$\,h for Stage~1, $12$\,h for Stage~2, and
$16$--$18$\,h for Stage~3. Code is available at
\url{https://anonymous.4open.science/r/arabic-LLM-legal-compression-6B5A}.

\subsection{Worked Examples}
\label{app:examples}

\paragraph{Answer-in-context (MCQ).}

\noindent
\foreignlanguage{arabic}{السؤال}:
\foreignlanguage{arabic}{
ما الذي لا يجوز اشتراطه في عقد المزارعة بخصوص نصيب أحد المتعاقدين؟
}

\noindent
\emph{Question:} What may not be stipulated in a sharecropping contract regarding
the share of one of the contracting parties?

\medskip

\noindent
\foreignlanguage{arabic}{الخيارات}:

\noindent
A) \foreignlanguage{arabic}{أن يكون مقداراً شائعاً من الناتج}
That it be an undivided proportion of the produce.

\noindent
B) \foreignlanguage{arabic}{أن يكون مقداراً غير شائع من الناتج أو ناتج موضع معين من الأرض أو الشجر}
That it be a non-proportional amount of the produce, or the produce of a
specific part of the land or trees.

\noindent
C) \foreignlanguage{arabic}{أن يكون ناتج الأرض بأكملها}
That it be the entire produce of the land.

\noindent
D) \foreignlanguage{arabic}{أن يكون ناتج الشجر بأكمله}
That it be the entire produce of the trees.

\medskip

\noindent
Correct answer: B

\medskip

\noindent
\foreignlanguage{arabic}{السياق}:
\foreignlanguage{arabic}{
لا يصح اشتراط أن يكون نصيب أحد المتعاقدين مقداراً غير شائع من الناتج أو ناتج موضع معين من الأرض أو الشجر.
}

\noindent
\emph{Context:} It is not valid to stipulate that the share of one of the contracting
parties be a non-proportional amount of the produce, or the produce of a specific
part of the land or trees.

\medskip

\noindent
\foreignlanguage{arabic}{نص الإجابة الصحيحة}:
\foreignlanguage{arabic}{
أن يكون مقداراً غير شائع من الناتج أو ناتج موضع معين من الأرض أو الشجر
}

\noindent
\emph{Correct answer text:} A non-proportional amount of the produce, or the produce
of a specific part of the land or trees.

\paragraph{Question (QA) Example}
\noindent
\foreignlanguage{arabic}{السؤال}:
\foreignlanguage{arabic}{
هل يمكن إفراغ عقار هبة لأكثر من شخص عن طريق منصة البورصة العقارية؟
}

\noindent
\emph{Question:} Can a gifted property be transferred to more than one person through the Real Estate Market Platform?

\medskip

\noindent
\foreignlanguage{arabic}{الإجابة}:
\foreignlanguage{arabic}{
نعم، إمكانية وهب عقار لأكثر من شخص متاحة في منصة البورصة العقارية.
}

\noindent
\emph{Answer:} Yes. The Real Estate Market Platform allows a property to be gifted to more than one person.

\section{Additional Results}
\label{sec:add_result}

\paragraph{Answer-in-context analysis.}
We further examined with-context MCQ performance according to whether the correct option text appeared verbatim in the associated legal passage, using the FANAR-9B INT4 NF4 model from Stage~2 (Table~\ref{tab:s2}). Of the 500 evaluation examples, 270 (54.0\%) contained the correct answer text, while 230 (46.0\%) did not. The model answered 268/270 (99.3\%) correctly in the former group and 227/230 (98.7\%) in the latter, for an overall accuracy of 99.0\% (495/500). The small observed difference suggests that exact answer-text overlap alone does not account for the model's high with-context accuracy. However, this analysis does not distinguish contextual inference from reliance on previously acquired knowledge.

\subsection{Detailed Stage~1 Results}
\label{app:stage1_detail}

Table~\ref{tab:stage1_all_models} presents the
complete Stage~1 results for FANAR, JAIS, and
Qwen across four adaptation configurations.
We report MCQ accuracy and correct predictions
on the fixed 500-item evaluation subset,
together with ROUGE-1 and ROUGE-L on the
200-item QA set under both evaluation protocols.
These results supplement the comparisons
in Section~\ref{sec:results_stage1}.


\begin{table*}[t]
\centering
\small
\setlength{\tabcolsep}{4pt}
\begin{tabular}{lllrrrr}
\toprule
\textbf{Model} &
\textbf{Setting} &
\textbf{Experiment} &
\textbf{MCQ Accuracy (\%)} &
\textbf{MCQ Correct} &
\textbf{QA ROUGE-1} &
\textbf{QA ROUGE-L} \\
\midrule

\multirow{8}{*}{FANAR}
& \multirow{4}{*}{With-Ctx}
& Exp A (Base)         & 97.8 & 489/500 & 0.0083 & 0.0083 \\
&& Exp B (CPT)         & 42.6 & 213/500 & 0.1700 & 0.1700 \\
&& Exp C (SFT)         & \textbf{98.8} & \textbf{494/500} & 0.1700 & 0.1700 \\
&& Exp D (CPT$\to$SFT) & \textbf{98.8} & \textbf{494/500} & 0.1683 & 0.1683 \\
\cmidrule(lr){2-7}
& \multirow{4}{*}{No-Ctx}
& Exp A (Base)         & 70.8 & 354/500 & 0.0058 & 0.0058 \\
&& Exp B (CPT)         & 74.8 & 374/500 & 0.1450 & 0.1450 \\
&& Exp C (SFT)         & 77.2 & 386/500 & 0.0490 & 0.0490 \\
&& Exp D (CPT$\to$SFT) & \textbf{79.2} & \textbf{396/500} & 0.0637 & 0.0637 \\
\midrule

\multirow{8}{*}{JAIS}
& \multirow{4}{*}{With-Ctx}
& Exp A (Base)         & 79.2 & 396/500 & 0.0194 & 0.0194 \\
&& Exp B (CPT)         & 93.6 & 468/500 & 0.0249 & 0.0249 \\
&& Exp C (SFT)         & 95.8 & 479/500 & 0.0001 & 0.0001 \\
&& Exp D (CPT$\to$SFT) & \textbf{98.6} & \textbf{493/500} & 0.0012 & 0.0012 \\
\cmidrule(lr){2-7}
& \multirow{4}{*}{No-Ctx}
& Exp A (Base)         & 71.6 & 358/500 & 0.0179 & 0.0179 \\
&& Exp B (CPT)         & 61.2 & 306/500 & 0.0008 & 0.0008 \\
&& Exp C (SFT)         & \textbf{76.4} & \textbf{382/500} & 0.0009 & 0.0009 \\
&& Exp D (CPT$\to$SFT) & 75.8 & 379/500 & 0.0012 & 0.0012 \\
\midrule

\multirow{8}{*}{Qwen}
& \multirow{4}{*}{With-Ctx}
& Exp A (Base)         & 97.4 & 487/500 & 0.0095 & 0.0095 \\
&& Exp B (CPT)         & 97.6 & 488/500 & 0.1532 & 0.1532 \\
&& Exp C (SFT)         & 98.6 & 493/500 & 0.3170 & 0.3170 \\
&& Exp D (CPT$\to$SFT) & \textbf{98.8} & \textbf{494/500} & 0.2750 & 0.2750 \\
\cmidrule(lr){2-7}
& \multirow{4}{*}{No-Ctx}
& Exp A (Base)         & 76.0 & 380/500 & 0.0058 & 0.0058 \\
&& Exp B (CPT)         & 78.4 & 392/500 & 0.2152 & 0.2152 \\
&& Exp C (SFT)         & 78.2 & 391/500 & 0.2950 & 0.2950 \\
&& Exp D (CPT$\to$SFT) & \textbf{79.8} & \textbf{399/500} & 0.2637 & 0.2637 \\

\bottomrule
\end{tabular}

\caption{Stage~1 results for FANAR, JAIS, and Qwen under with-context and no-context evaluation. MCQ results are reported on 500 items; bold denotes the best-performing variant within each model and setting.}
\label{tab:stage1_all_models}
\end{table*}